\documentclass[11pt,a4paper]{article}
\usepackage[nohyperref]{acl2017}
\usepackage{times}
\usepackage{latexsym}
\usepackage{color}

\usepackage{url}

\usepackage{amsmath}
\usepackage{graphicx}
\graphicspath{ {figs/} }
\usepackage{float}
\usepackage{calc}
\usepackage{multirow}

\aclfinalcopy 
\def\aclpaperid{590} 

\usepackage[font=small]{caption}

\title{Multi-Task Video Captioning with Video and Entailment Generation}

\author{Ramakanth Pasunuru \and Mohit Bansal \\
  UNC Chapel Hill \\
  {\tt \{ram, mbansal\}@cs.unc.edu} \\
 }

\date{}

\begin{document}
\maketitle
\begin{abstract}
Video captioning, the task of describing the content of a video, has seen some promising improvements in recent years with sequence-to-sequence models, but accurately learning the temporal and logical dynamics involved in the task still remains a challenge, especially given the lack of sufficient annotated data. We improve video captioning by sharing knowledge with two related directed-generation tasks: a temporally-directed unsupervised video prediction task to learn richer context-aware video encoder representations, and a logically-directed language entailment generation task to learn better video-entailed caption decoder representations. For this, we present a many-to-many multi-task learning model that shares parameters across the encoders and decoders of the three tasks. We achieve significant improvements and the new state-of-the-art on several standard video captioning datasets using diverse automatic and human evaluations. We also show mutual multi-task improvements on the entailment generation task.

\end{abstract}
\section{Introduction}
\label{sec-intro}

\begin{figure}
\centering
\includegraphics[width=0.98\linewidth]{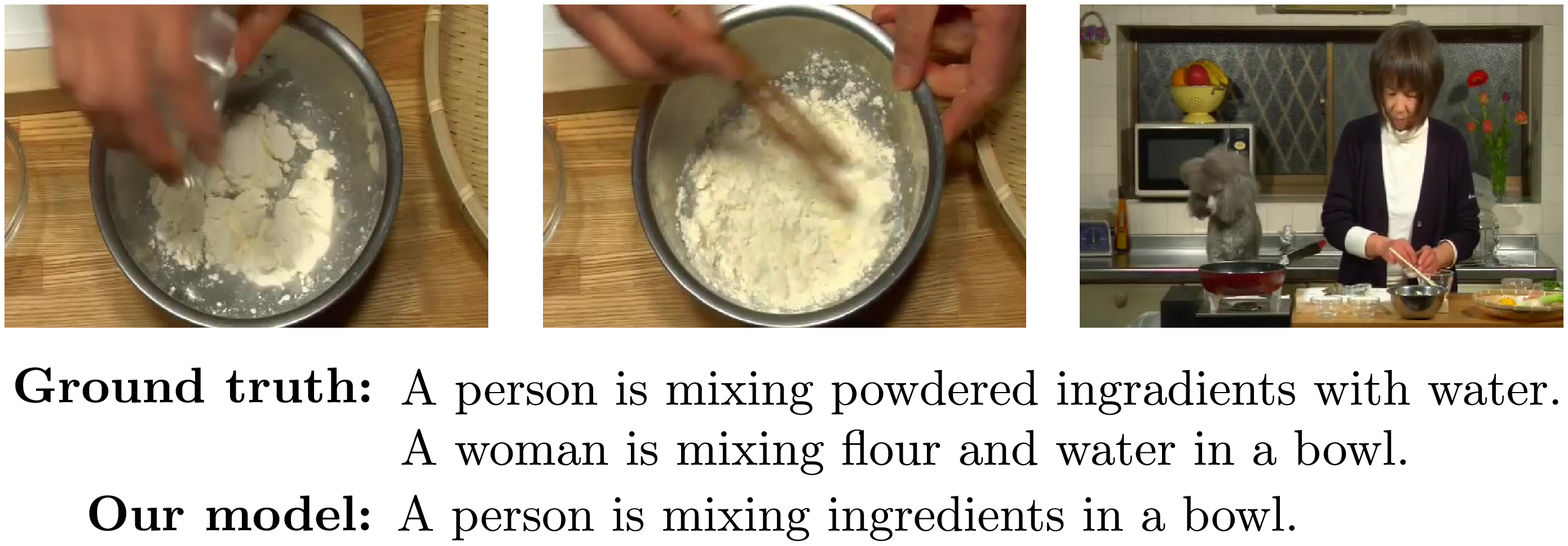}
\vspace{-3pt}
\caption{A video captioning example from the YouTube2Text dataset, with the ground truth captions and our many-to-many multi-task model's predicted caption.}
\vspace{-10pt}
\label{fig:introexample}
\end{figure}

Video captioning is the task of automatically generating a natural language description of the content of a video, as shown in Fig.~\ref{fig:introexample}. It has various applications such as assistance to a visually impaired person and improving the quality of online video search or retrieval.
This task has gained recent momentum in the natural language processing and computer vision communities, esp. with the advent of powerful image processing features as well as sequence-to-sequence LSTM models. It is also a step forward from static image captioning, because in addition to modeling the spatial visual features, the model also needs to learn the temporal across-frame action dynamics and the logical storyline language dynamics.

Previous work in video captioning \cite{venugopalan2015sequence,pan2015jointly} has shown that recurrent neural networks (RNNs) are a good choice for modeling the temporal information in the video. A sequence-to-sequence model is then used to `translate' the video to a caption. \newcite{venugopalan2016improving} showed linguistic improvements over this by fusing the decoder with external language models. Furthermore, an attention mechanism between the video frames and the caption words captures some of the temporal matching relations better~\cite{yao2015describing,pan2015hierarchical}.
More recently, hierarchical two-level RNNs were proposed to allow for longer inputs and to model the full paragraph caption dynamics of long video clips~\cite{pan2015hierarchical,yu2015video}. 

Despite these recent improvements, video captioning models still suffer from the lack of sufficient temporal and logical supervision to be able to correctly capture the action sequence and story-dynamic language in videos, esp. in the case of short clips. Hence, they would benefit from incorporating such complementary directed knowledge, both visual and textual.
We address this by jointly training the task of video captioning with two related directed-generation tasks: a temporally-directed unsupervised video prediction task and a logically-directed language entailment generation task.
We model this via many-to-many multi-task learning based sequence-to-sequence models~\cite{luong2015multi} that allow the sharing of parameters among the encoders and decoders across the three different tasks, with additional shareable attention mechanisms.

The unsupervised video prediction task, i.e., video-to-video generation (adapted from ~\newcite{srivastava2015unsupervised}), shares its encoder with the video captioning task's encoder, and helps it learn richer video representations that can predict their temporal context and action sequence. The entailment generation task, i.e., premise-to-entailment generation (based on the image caption domain SNLI corpus~\cite{bowman2015large}), shares its decoder with the video captioning decoder, and helps it learn better video-entailed caption representations, since the caption is essentially an entailment of the video, i.e., it describes subsets of objects and events that are logically implied by (or follow from) the full video content. The overall many-to-many multi-task model combines all three tasks.

Our three novel multi-task models show statistically significant improvements over the state-of-the-art, and achieve the best-reported results (and rank) on multiple datasets, based on several automatic and human evaluations. We also demonstrate that video captioning, in turn, gives mutual improvements on the new multi-reference entailment generation task.

\section{Related Work}
\label{sec-relatedwork}
Early video captioning work~\cite{guadarrama2013youtube2text,thomason2014integrating,huang2012multi} used a two-stage pipeline to first extract a subject, verb, and object (S,V,O) triple and then generate a sentence based on it. \newcite{venugopalan2014translating} fed mean-pooled static frame-level visual features (from convolution neural networks pre-trained on image recognition) of the video as input to the language decoder. To harness the important frame sequence temporal ordering,~\newcite{venugopalan2015sequence} proposed a sequence-to-sequence model with video encoder and language decoder RNNs. 

More recently,~\newcite{venugopalan2016improving} explored linguistic improvements to the caption decoder by fusing it with external language models. 
Moreover, an attention or alignment mechanism was added between the encoder and the decoder to learn the temporal relations (matching) between the video frames and the caption words~\cite{yao2015describing,pan2015hierarchical}.
In contrast to static visual features, \newcite{yao2015describing} also considered temporal video features from a 3D-CNN model pre-trained on an action recognition task.

To explore long range temporal relations,~\newcite{pan2015hierarchical} proposed a two-level hierarchical RNN encoder which limits the length of input information and allows temporal transitions between segments.~\newcite{yu2015video}'s hierarchical RNN generates sentences at the first level and the second level captures inter-sentence dependencies in a paragraph. \newcite{pan2015jointly} proposed to simultaneously learn the RNN word probabilities and a visual-semantic joint embedding space that enforces the relationship between the semantics
of the entire sentence and the visual content.
Despite these useful recent improvements, video captioning still suffers from limited supervision and generalization capabilities, esp. given the complex action-based temporal and story-based logical dynamics that need to be captured from short video clips. Our work addresses this issue by bringing in complementary temporal and logical knowledge from video prediction and textual entailment generation tasks (respectively), and training them together via many-to-many multi-task learning.

Multi-task learning is a useful learning paradigm to improve the supervision and the generalization performance of a task by jointly training it with related tasks~\cite{caruana1998multitask,argyriou2007multi,kumar2012learning}. Recently,~\newcite{luong2015multi} combined multi-task learning with sequence-to-sequence models, sharing parameters across the tasks' encoders and decoders. They showed improvements on machine translation using parsing and image captioning. We additionally incorporate an attention mechanism to this many-to-many multi-task learning approach and improve the multimodal, temporal-logical video captioning task by sharing its video encoder with the encoder of a video-to-video prediction task and by sharing its caption decoder with the decoder of a linguistic premise-to-entailment generation task.

Image representation learning has been successful via supervision from very large object-labeled datasets. However, similar amounts of supervision are lacking for video representation learning. ~\newcite{srivastava2015unsupervised} address this by proposing unsupervised video representation learning  via sequence-to-sequence RNN models, where they reconstruct the input video sequence or predict the future sequence. We model video generation with an attention-enhanced encoder-decoder and harness it to improve video captioning.

The task of recognizing textual entailment (RTE) is to classify whether the relationship between a premise and hypothesis sentence is that of entailment (i.e., logically follows), contradiction, or independence (neutral), which is helpful for several downstream NLP tasks. The recent Stanford Natural Language Inference (SNLI) corpus by~\newcite{bowman2015large} allowed training end-to-end neural networks that outperform earlier feature-based RTE models~\cite{lai2014illinois,jimenez2014unal}. However, directly generating the entailed hypothesis sentences given a premise sentence would be even more beneficial than retrieving or reranking sentence pairs, because most downstream generation tasks only come with the source sentence and not pairs. Recently,~\newcite{kolesnyk2016generating} tried a  sequence-to-sequence model for this on the original SNLI dataset, which is a single-reference setting and hence restricts automatic evaluation. We modify the SNLI corpus to a new multi-reference (and a more challenging zero train-test premise overlap) setting, and present a novel multi-task training setup with the related video captioning task (where the caption is also entailed by the video), showing mutual improvements on both the tasks.


\begin{figure}
\centering
\includegraphics[width=0.98\linewidth]{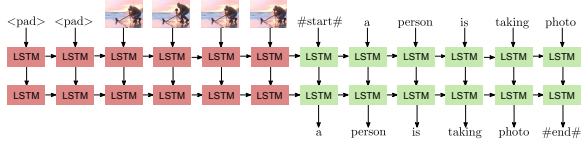}
\vspace{-3pt}
\caption{Baseline sequence-to-sequence model for video captioning: standard encoder-decoder LSTM-RNN model.}
\vspace{-10pt}
\label{fig:basemodel}
\end{figure}

\section{Models}
\label{sec-models}
We first discuss a simple encoder-decoder model as a baseline reference for video captioning. Next, we improve this via an attention mechanism. Finally, we present similar models for the unsupervised video prediction and  entailment generation tasks, and then combine them with video captioning via the many-to-many multi-task approach.

\begin{figure}
\centering
\includegraphics[width=0.75\linewidth]{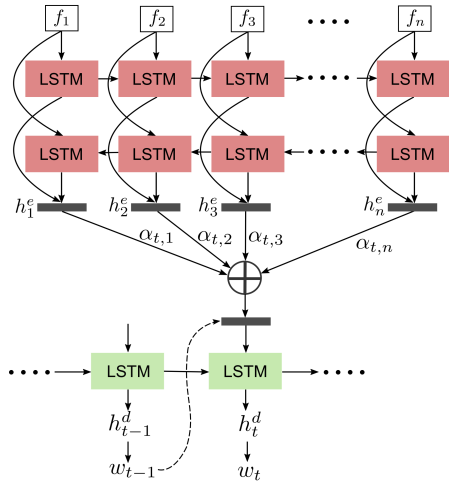}
\vspace{-3pt}
\caption{Attention-based sequence-to-sequence baseline model for video captioning (similar models also used for video prediction and entailment generation).}
\vspace{-10pt}
\label{fig:attention}
\end{figure}

\subsection{Baseline Sequence-to-Sequence Model}
\label{subsec-basemodel}
Our baseline model is similar to the standard machine translation encoder-decoder RNN model~\cite{sutskever2014sequence} where the final state of the encoder RNN is input as an initial state to the decoder RNN, as shown in Fig.~\ref{fig:basemodel}. The RNN is based on Long Short Term Memory (LSTM) units, which are good at memorizing long sequences due to forget-style gates~\cite{hochreiter1997long}. For video captioning, our input to the encoder is the video frame features\footnote{We use several popular image features such as VGGNet, GoogLeNet and Inception-v4. Details in Sec.~\ref{subsec-datasets}.}  $\{f_1,f_2,...,f_n\}$ of length $n$, and the caption word sequence $\{w_1,w_2,...,w_m\}$ of length $m$ is generated during the decoding phase. The distribution  of the output sequence w.r.t. the input sequence is:
\begin{equation}
p(w_1,...,w_m|f_1,...,f_n) =  \prod_{t=1}^{m} p(w_t|h^d_{t})
\end{equation}
where $h^d_t$ is the hidden state at the $t^{th}$ time step of the decoder RNN, obtained from $h_{t-1}^d$ and $w_{t-1}$ via the standard LSTM-RNN equations. The distribution $p(w_t|h_t^d)$ is given by \emph{softmax} over all the words in the vocabulary.

\begin{figure*}
\centering
\includegraphics[width=0.96\linewidth]{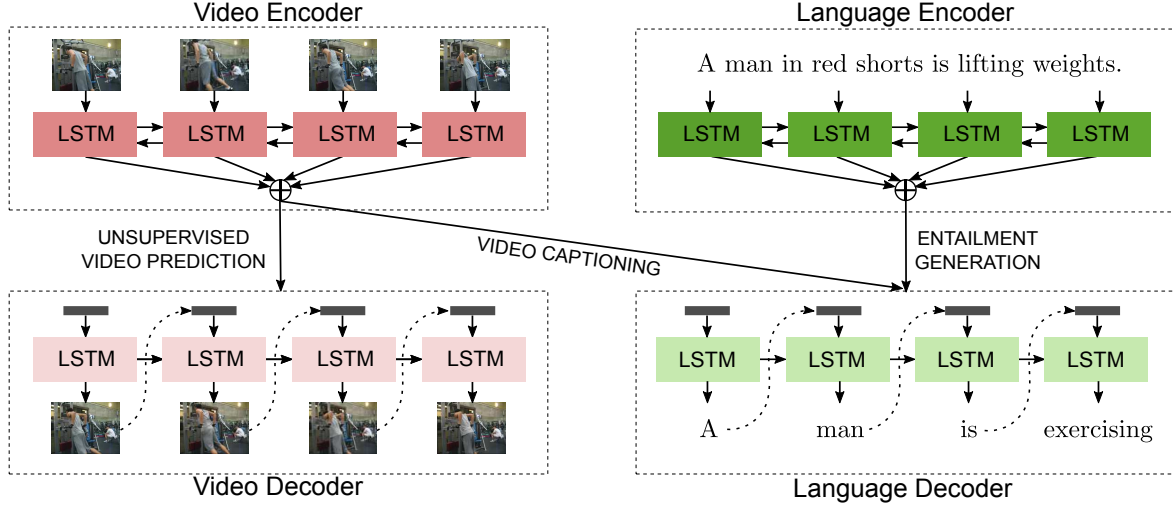}
\vspace{-6pt}
\caption{Our many-to-many multi-task learning model to share encoders and decoders of the video captioning, unsupervised video prediction, and entailment generation tasks.}
\vspace{-10pt}
\label{fig-multitask}
\end{figure*}

\subsection{Attention-based Model}
\label{subsec-atttentionmodel}
Our attention model architecture is similar to~\newcite{bahdanau2014neural}, with a bidirectional LSTM-RNN as the encoder and a unidirectional LSTM-RNN as the decoder, see Fig.~\ref{fig:attention}.  
At each time step $t$, the decoder LSTM hidden state $h^d_t$ is a non-linear recurrent function of the previous decoder hidden state $h^d_{t-1}$, the previous time-step's generated word $w_{t-1}$, and the context vector $c_t$:
\begin{equation}
\label{eqn-decoder}
h^d_t = S(h^d_{t-1}, w_{t-1}, c_t)
\end{equation}
where $c_t$ is a weighted sum of encoder hidden states $\{h^e_i\}$:
\begin{equation}
c_t = \sum_{i=1}^{n} \alpha_{t,i} h^e_i
\end{equation}
These attention weights \{$\alpha_{t,i}$\} act as an alignment mechanism by giving higher weights to certain encoder hidden states which match that decoder time step better, and are computed as:
\begin{equation}
\alpha_{t,i} =  \frac{exp(e_{t,i})}{\sum_{k=1}^n exp(e_{t,k})}
\end{equation}
where the attention function $e_{t,i}$ is defined as:
\begin{equation}
e_{t,i} =  w^T tanh(W^e_a h^e_i + W^d_a h^d_{t-1}+b_a)
\end{equation}
where $w$, $W^e_a$, $W^d_a$, and $b_a$ are learned parameters.
This attention-based sequence-to-sequence model (Fig.~\ref{fig:attention}) is our enhanced baseline for video captioning. We next discuss similar models for the new tasks of unsupervised video prediction and entailment generation and then finally share them via multi-task learning.

\subsection{Unsupervised Video Prediction}
\label{sec-videopred}

We model unsupervised video representation by predicting the sequence of future video frames given the current frame sequence. Similar to Sec.~\ref{subsec-atttentionmodel}, a bidirectional LSTM-RNN encoder and an LSTM-RNN decoder is used, along with attention. If the frame level features of a video of length $n$ are $\{f_1,f_2,...,f_n\}$, these are divided into two sets such that given the current frames $\{f_1,f_2,..,f_k\}$ (in its encoder), the model has to predict (decode) the rest of the frames $\{f_{k+1},f_{k+2},..,f_n\}$. The motivation is that this helps the video encoder learn rich temporal representations that are aware of their action-based context and are also robust to missing frames and varying frame lengths or motion speeds.
The optimization function is defined as:
\begin{equation}
\underset{\phi}{\text{minimize}}  \sum_{t=1}^{n-k}||f_t^d-f_{t+k}||_2^2
\end{equation}
where  $\phi$ are the model parameters,  $f_{t+k}$ is the true future frame feature at decoder time step $t$ and $f_t^d$ is the decoder's predicted future frame feature at decoder time step $t$, defined as:
\begin{equation}
f_t^d =  S(h_{t-1}^d,f_{t-1}^d,c_t)
\end{equation}
similar to Eqn.~\ref{eqn-decoder}, with $h_{t-1}^d$ and $f_{t-1}^d$ as the previous time step's hidden state and predicted frame feature respectively, and $c_t$ as the attention-weighted context vector.

\subsection{Entailment Generation}
\label{subsec-entailmentpredictionmodel}
Given a sentence (premise), the task of entailment generation is to generate a sentence (hypothesis) which is a logical deduction or implication of the premise. Our entailment generation model again uses a bidirectional LSTM-RNN encoder and LSTM-RNN decoder with an attention mechanism (similar to Sec.~\ref{subsec-atttentionmodel}).
If the premise $s^p$ is a sequence of words $\{w_1^p,w_2^p,...,w_n^p\}$ and the hypothesis $s^h$ is $\{w_1^h,w_2^h,...,w_m^h\}$, the distribution of the entailed hypothesis w.r.t. the premise is:
\vspace{-4pt}
\begin{equation}
\vspace{-3pt}
p(w_1^h,...,w_m^h|w_1^p,...,w_n^p) =  \prod_{t=1}^{m} p(w_t^h|h^d_{t})
\end{equation}
where the distribution $p(w_t^h|h_t^d)$ is again obtained via softmax over all the words in the vocabulary and the decoder state $h_t^d$ is similar to Eqn.~\ref{eqn-decoder}.

\subsection{Multi-Task Learning}
Multi-task learning helps in sharing information between different tasks and across domains. Our primary aim is to improve the video captioning model, where visual content translates to a textual form in a directed (entailed) generation way. Hence, this presents an interesting opportunity to share temporally and logically directed knowledge with both visual and linguistic generation tasks.  Fig.~\ref{fig-multitask} shows our overall many-to-many multi-task model for jointly learning video captioning, unsupervised video prediction, and textual entailment generation. Here, the video captioning task shares its video encoder (parameters) with the encoder of the video prediction task (one-to-many setting) so as to learn context-aware and temporally-directed visual representations (see Sec.~\ref{sec-videopred}). 

Moreover, the decoder of the video captioning task is shared with the decoder of the textual entailment generation task (many-to-one setting), thus helping generate captions that can be `entailed' by, i.e., are logically implied by or follow from the video content (see Sec.~\ref{subsec-entailmentpredictionmodel}).\footnote{Empirically, logical entailment helped captioning more than simple fusion with language modeling (i.e., partial sentence completion with no logical implication), because a caption is also `entailed' by a video in a logically-directed sense and hence the entailment generation task matches the video captioning task better than language modeling. Moreover, a multi-task setup is more suitable to add directed information such as entailment (as opposed to pretraining or fusion with only the decoder). Details in Sec.~\ref{youtube-results}.} In both the one-to-many and the many-to-one settings, we also allow the attention parameters to be shared or separated. The overall many-to-many setting thus improves both the visual and language representations of the video captioning model. 

We train the multi-task model by alternately optimizing each task in mini-batches based on a mixing ratio. Let $\alpha_v$, $\alpha_f$, and $\alpha_e$ be the  number of mini-batches optimized alternately from each of these three tasks -- video captioning, unsupervised video future frames prediction, and entailment generation, resp.  Then the mixing ratio is defined as  $\frac{\alpha_v}{(\alpha_v+\alpha_f+\alpha_e)}:\frac{\alpha_f}{(\alpha_v+\alpha_f+\alpha_e)}:\frac{\alpha_e}{(\alpha_v+\alpha_f+\alpha_e)}$.

\section{Experimental Setup}
\label{setup}

\subsection{Datasets}
\label{subsec-datasets}

\paragraph{Video Captioning Datasets}
We report results on three popular video captioning datasets. First, we use the YouTube2Text or MSVD~\cite{chen2011collecting} for our primary results, which contains $1970$ YouTube videos in the wild with several different reference captions per video ($40$ on average).  We also use MSR-VTT~\cite{xu2016msr} with $10,000$ diverse video clips (from a video search engine) -- it has  $200,000$ video clip-sentence pairs and around $20$ captions per video; and M-VAD~\cite{torabi2015using} with $49,000$ movie-based video clips but only $1$ or $2$ captions per video, making most evaluation metrics (except paraphrase-based METEOR) infeasible.
We use the standard splits for all three datasets.
Further details about all these datasets are provided in the supplementary.

\paragraph{Video Prediction Dataset}
For our unsupervised video representation learning task, we use the UCF-101 action videos dataset~\cite{soomro2012ucf101}, which contains $13,320$ video clips of $101$ action categories, and
suits our video captioning task well because it also contains short video clips of a single action or few actions. We use the standard splits -- further details in supplementary.

\paragraph{Entailment Generation Dataset}
For the entailment generation encoder-decoder model, we use the Stanford Natural Language Inference (SNLI) corpus~\cite{bowman2015large}, which contains human-annotated English sentence pairs with classification labels of entailment, contradiction and neutral. It has a total of $570,152$ sentence pairs out of which $190,113$ correspond to true entailment pairs, and we use this subset in our multi-task video captioning model.  For improving video captioning, we use the same training/validation/test splits as provided by~\newcite{bowman2015large}, which is $183,416$ training, $3,329$ validation, and $3,368$ testing pairs (for the entailment subset). 

However, for the entailment generation multi-task results (see results in~Sec.~\ref{subsec-entailmentprediction}), we modify the splits so as to create a multi-reference setup which can afford evaluation with automatic metrics. A given premise usually has multiple entailed hypotheses but the original SNLI corpus is set up as single-reference (for classification). Due to this, the different entailed hypotheses of the same premise land up in different splits of the dataset (e.g., one in train and one in test/validation) in many cases. Therefore, we regroup the premise-entailment pairs and modify the split as follows:  among the $190,113$ premise-entailment pairs subset of the SNLI corpus, there are $155,898$ unique premises; out of which $145,822$ have only one hypothesis and we make this the training set, and the rest of them ($10,076$) have more than one hypothesis, which we randomly shuffle and divide equally into test and validation sets, so that each of these two sets has approximately the same distribution of the number of reference hypotheses per premise. 

These new validation and test sets hence contain premises with multiple entailed hypotheses as ground truth references, thus allowing for automatic metric evaluation, where differing generations still get positive scores by matching one of the multiple references. Also, this creates a more challenging dataset for entailment generation because of zero premise overlap between the training and val/test sets. We will make these split details publicly available.

\paragraph{Pre-trained Visual Frame Features}
For the three video captioning and UCF-101 datasets, we fix our sampling rate to $3fps$ to bring uniformity in the temporal representation of actions across all videos. These sampled frames are then converted into features using several state-of-the-art pre-trained models on ImageNet~\cite{deng2009imagenet} -- VGGNet~\cite{simonyan2014very}, GoogLeNet~\cite{szegedy2015going,ioffe2015batch}, and Inception-v4~\cite{szegedy2016inception}. Details of these feature dimensions and layer positions are in the supplementary.

\subsection{Evaluation (Automatic and Human)}
\label{subsec-evaluationmetrics}
For our video captioning as well as entailment generation results, we use four diverse automatic evaluation metrics that are popular for image/video captioning and language generation in general: METEOR~\cite{banerjee2005meteor}, BLEU-4~\cite{papineni2002bleu}, CIDEr-D~\cite{vedantam2015cider}, and ROUGE-L~\cite{lin2004rouge}. Particularly, METEOR and CIDEr-D have been justified to be better for generation tasks, because CIDEr-D uses consensus among the (large) number of references and METEOR uses soft matching based on stemming, paraphrasing, and WordNet synonyms. We use the standard evaluation code from the Microsoft COCO server~\cite{chen2015microsoft} to obtain these results and also to compare the results with previous papers.\footnote{We use avg. of these four metrics on validation set to choose the best model, except for single-reference M-VAD dataset where we only report and choose based on METEOR.}

We also present human evaluation results based on relevance (i.e., how related is the generated caption w.r.t. the video contents such as actions, objects, and events; or is the generated hypothesis entailed or implied by the premise) and coherence (i.e., a score on the logic, readability, and fluency of the generated sentence).

\subsection{Training Details}
\label{sec:trainingdetails}
We tune all hyperparameters on the dev splits: LSTM-RNN hidden state size, learning rate, weight initializations, and mini-batch mixing ratios (tuning ranges in supplementary). We use the following settings in all of our models (unless otherwise specified): we unroll video encoder/decoder RNNs to $50$ time steps and language encoder/decoder RNNs to $30$ time steps. We use a 1024-dimension RNN hidden state size and $512$-dim vectors to embed visual features and word vectors. We use Adam optimizer~\cite{kingma2014adam}. We apply a dropout of $0.5$. See subsections below and supp for full details.

\begin{table*}
\begin{center}
\begin{small}
\begin{tabular}{|l|c|c|c|c|}
\hline
Models & METEOR & CIDEr-D & ROUGE-L & BLEU-4 \\
\hline
\multicolumn{5}{|c|}{\textsc{Previous Work}}\\
\hline
LSTM-YT ({\small V}) \cite{venugopalan2014translating} & 26.9 & - & - & 31.2 \\

S2VT ({\small V + A}) \cite{venugopalan2015sequence} & 29.8 & - & - & - \\

Temporal Attention ({\small G + C}) \cite{yao2015describing} & 29.6 & 51.7 & - & 41.9 \\
LSTM-E ({\small V + C}) \cite{pan2015jointly} & 31.0 & - & - & 45.3 \\
Glove + DeepFusion ({\small V}) ({\small E}) \cite{venugopalan2016improving}& 31.4 & - & - & 42.1 \\ 
p-RNN ({\small V + C}) \cite{yu2015video} & 32.6 & 65.8 & - & 49.9 \\
HNRE + Attention ({\small G + C}) \cite{pan2015hierarchical} & 33.9 & - & - & 46.7 \\
\hline
\multicolumn{5}{|c|}{\textsc{Our Baselines}}\\
\hline
Baseline ({\small V}) & 31.4 & 63.9 & 68.0 & 43.6 \\
Baseline ({\small G}) & 31.7 & 64.8 & 68.6 & 44.1  \\
Baseline ({\small I}) & 33.3 & 75.6 & 69.7 & 46.3  \\
Baseline + Attention ({\small V}) & 32.6 & 72.2 & 69.0 & 47.5 \\
Baseline + Attention ({\small G}) & 33.0 & 69.4 & 68.3 & 44.9\\
Baseline + Attention ({\small I}) & 33.8 & 77.2 & 70.3 & 49.9\\
Baseline + Attention ({\small I}) ({\small E}) $\otimes$  & 35.0 & 84.4 & 71.5 & 52.6 \\
\hline
\multicolumn{5}{|c|}{\textsc{Our Multi-Task Learning Models}}\\
\hline
$\otimes$ + Video Prediction (1-to-M) & 35.6 & 88.1 & 72.9 & 54.1  \\
$\otimes$ + Entailment Generation (M-to-1)  & 35.9 & 88.0 & 72.7 & 54.4 \\
$\otimes$ + Video Prediction + Entailment Generation (M-to-M) \ \ \ \ \ \ \ & \textbf{36.0} & \textbf{92.4} & \textbf{72.8} & \textbf{54.5} \\
\hline
\end{tabular}
\end{small}
\end{center}
\vspace{-7pt}
\caption{Primary video captioning results on Youtube2Text (MSVD), showing previous works, our several strong baselines, and our three multi-task models. Here, V, G, I, C, A are short for VGGNet, GoogLeNet, Inception-v4, C3D, and AlexNet visual features; E = ensemble. The multi-task models are applied on top of our best video captioning baseline $\otimes$, with an ensemble. All the multi-task models are statistically significant over the baseline (discussed inline in the corresponding results sections).}
\vspace{-5pt}
\label{table-baseandmultitaskresults}
\end{table*}

\section{Results and Analysis}
\label{sec-results}

\subsection{Video Captioning on YouTube2Text}
\label{youtube-results}

Table~\ref{table-baseandmultitaskresults} presents our primary results on the YouTube2Text (MSVD) dataset, reporting several previous works, all our baselines and attention model ablations, and our three multi-task models, using the four automated evaluation metrics. For each subsection below, we have reported the important training details inline, and refer to the supplementary for full details (e.g., learning rates and initialization).

\paragraph{Baseline Performance} 
 
We first present all our baseline model choices (ablations) in Table~\ref{table-baseandmultitaskresults}. Our baselines represent the standard sequence-to-sequence model with three different visual feature types as well as those with attention mechanisms. Each baseline model is trained with three random seed initializations and the average is reported (for stable results). The final baseline model $\otimes$ instead uses an ensemble (E), which is a standard denoising method~\cite{sutskever2014sequence} that performs inference over ten randomly initialized models, i.e., at each time step $t$ of the decoder, we generate a word based on the avg. of the likelihood probabilities from the ten models. Moreover, we use beam search with size $5$ for all baseline models. Overall, the final baseline model with Inception-v4 features, attention, and 10-ensemble performs well (and is better than all previous state-of-the-art), and so we next add all our novel multi-task models on top of this final baseline.

\paragraph{Multi-Task with Video Prediction (1-to-M)} 
 
Here, the video captioning and unsupervised video prediction tasks share their encoder LSTM-RNN weights and image embeddings in a one-to-many multi-task setting. Two important hyperparameters tuned (on the validation set of captioning datasets) are the ratio of encoder vs decoder frames for video prediction on UCF-101 (where we found that $80\%$ of frames as input and $20\%$ for prediction performs best); and the mini-batch mixing ratio between the captioning and video prediction tasks (where we found $100:200$ works well). Table~\ref{table-baseandmultitaskresults} shows a statistically significant improvement\footnote{Statistical significance of $p < 0.01$ for CIDEr-D and ROUGE-L, $p < 0.02$ for BLEU-4, $p < 0.03$ for METEOR, based on the bootstrap test~\cite{noreen1989computer,efron1994introduction} with 100K samples.} in all metrics in comparison to the best baseline (non-multitask) model as well as w.r.t. all previous works, demonstrating the effectiveness of multi-task learning for video captioning with video prediction, even with unsupervised signals.

\paragraph{Multi-Task with Entailment Generation (M-to-1)} Here, the video captioning and entailment generation tasks share their language decoder LSTM-RNN weights and word embeddings in a many-to-one multi-task setting. We observe that a mixing ratio of $100:50$ alternating mini-batches (between the captioning and entailment tasks) works well here. Again, Table~\ref{table-baseandmultitaskresults} shows statistically significant improvements\footnote{Statistical significance of $p < 0.01$ for all four metrics.} in all the metrics in comparison to the best baseline model (and all previous works) under this multi-task setting. Note that in our initial experiments, our entailment generation model helped the video captioning task significantly more than the alternative approach of simply improving fluency by adding (or deep-fusing) an external language model (or pre-trained word embeddings) to the decoder (using both in-domain and out-of-domain language models), again because a caption is also `entailed' by a video in a logically-directed sense and hence this matches our captioning task better (also see results of~\newcite{venugopalan2016improving} in Table~\ref{table-baseandmultitaskresults}).

\paragraph{Multi-Task with Video and Entailment Generation (M-to-M)}

Combining the above one-to-many and many-to-one multi-task learning models, our full model is the 3-task, many-to-many model (Fig.~\ref{fig-multitask}) where both the video encoder and the language decoder of the video captioning model are shared (and hence improved) with that of the unsupervised video prediction and entailment generation models, respectively.\footnote{We found the setting with unshared attention parameters to work best, likely because video captioning and video prediction prefer very different alignment distributions.} A mixing ratio of $100:100:50$ alternate mini-batches of video captioning, unsupervised video prediction, and entailment generation, resp. works well. Table~\ref{table-baseandmultitaskresults} shows that our many-to-many multi-task model again outperforms our strongest baseline (with statistical significance of $p < 0.01$ on all metrics), as well as all the previous state-of-the-art results by large absolute margins on all metrics. It also achieves significant improvements on some metrics over the one-to-many and many-to-one models.\footnote{Many-to-many model's improvements have a statistical significance of $p < 0.01$ on all metrics w.r.t. baseline, and $p < 0.01$ on CIDEr-D w.r.t. both one-to-many and many-to-one models, and $p < 0.04$ on METEOR w.r.t. one-to-many.} Overall, we achieve the best results to date on YouTube2Text (MSVD) on all metrics.

\begin{figure*}
\centering
\includegraphics[width=1\textwidth]{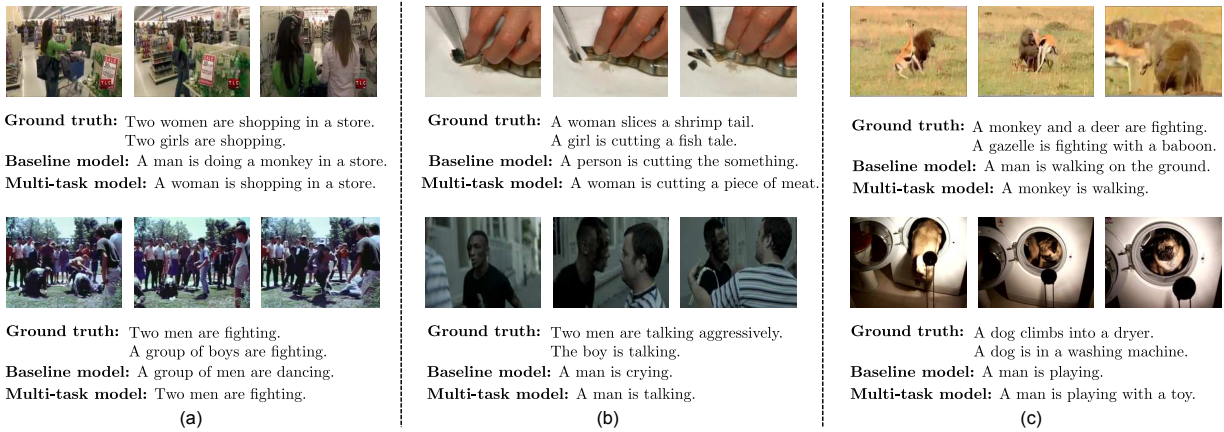}
\vspace{-17pt}
\caption{Examples of generated video captions on the YouTube2Text dataset: (a) complex examples where the multi-task model performs better than the baseline; (b) ambiguous examples (i.e., ground truth itself confusing) where multi-task model still correctly predicts one of the possible categories (c) complex examples where both models perform poorly.\vspace{-4pt}}
\label{fig-msvdexamples}
\end{figure*}


\begin{table}
\begin{small}
\begin{center}
\begin{tabular}{|l|c|c|c|c|}
\hline
Models & M & C & R & B\\
\hline
Venugopalan ~\shortcite{venugopalan2014translating}$^\star$ & 23.4 & - & - & 32.3\\
\newcite{yao2015describing}$^\star$ & 25.2 & - & - & 35.2 \\
\newcite{xu2016msr} & 25.9 & - & - & 36.6 \\
\hline
Rank1: v2t\_navigator & 28.2 & 44.8 &  \textbf{60.9} & \textbf{40.8} \\
Rank2: Aalto & 26.9 & 45.7 & 59.8 & 39.8\\
Rank3: VideoLAB & 27.7 & 44.1 & 60.6 & 39.1 \\
\hline
Our Model (\textbf{New Rank1}) & \textbf{28.8} & \textbf{47.1} & 60.2 & \textbf{40.8} \\
\hline
\end{tabular}
\vspace{-4pt}
\caption{Results on MSR-VTT dataset on the 4 metrics. $^\star$Results are reimplementations as per~\newcite{xu2016msr}. We also report the top 3 leaderboard systems -- our model achieves the new rank 1 based on their ranking method.}
\vspace{-5pt}
\label{table-msrvttresults}
\end{center}
\end{small}
\end{table}

\begin{table}
\begin{center}
\begin{small}
\begin{tabular}{|l|c|}
\hline
Models  & METEOR \\
\hline
\newcite{yao2015describing} & 5.7 \\
\newcite{venugopalan2015sequence}& 6.7 \\
\newcite{pan2015hierarchical} & 6.8 \\

\hline
Our M-to-M Multi-Task Model & \textbf{7.4}\\
\hline
\end{tabular}
\vspace{-5pt}
\caption{Results on M-VAD dataset.}
\vspace{-12pt}
\label{table-mvadresults}
\end{small}
\end{center}
\end{table}

\subsection{Video Captioning on MSR-VTT, M-VAD}
\label{msrvttmvad-results}
In Table~\ref{table-msrvttresults}, we also train and evaluate our final many-to-many multi-task model on two other video captioning datasets (using their standard splits; details in supplementary). First, we evaluate on the new MSR-VTT dataset~\cite{xu2016msr}.  Since this is a recent dataset, we list previous works' results as reported by the MSR-VTT dataset paper itself.\footnote{In their updated supplementary at \scriptsize{\url{https://www.microsoft.com/en-us/research/wp-content/uploads/2016/10/cvpr16.supplementary.pdf}}} 
We improve over all of these significantly. Moreover, they maintain a leaderboard\footnote{\scriptsize{\url{http://ms-multimedia-challenge.com/leaderboard}}} on this dataset and we also report the top 3 systems from it. Based on their ranking method, our multi-task model achieves the new rank 1 on this leaderboard. 
In Table~\ref{table-mvadresults}, we further evaluate our model on the challenging movie-based M-VAD dataset, and again achieve improvements over all previous work ~\cite{venugopalan2015sequence,pan2015hierarchical,yao2015describing}.\footnote{Following previous work, we only use METEOR because M-VAD only has a single reference caption per video.}

\begin{table}[t]
\begin{center}
\begin{small}
\begin{tabular}{|l|c|c|c|c|}
\hline
Models & M & C & R & B \\
\hline
Entailment Generation & 29.6 & 117.8& 62.4 & 40.6 \\
+Video Caption (M-to-1) & \textbf{30.0} & \textbf{121.6} & \textbf{63.9} & \textbf{41.6} \\
\hline
\end{tabular}
\vspace{-10pt}
\end{small}
\end{center}
\caption{Entailment generation results with the four metrics.}
\vspace{-10pt}
\label{table-entailmentprediction}
\end{table}


\subsection{Entailment Generation Results}
\label{subsec-entailmentprediction}
Above, we showed that the new entailment generation task helps improve video captioning. Next, we show that the video captioning task also inversely helps the entailment generation task. Given a premise, the task of entailment generation is to generate an entailed hypothesis. We use only the entailment pairs subset of the SNLI corpus for this, but with a multi-reference split setup to allow automatic metric evaluation and a zero train-test premise overlap (see Sec.~\ref{subsec-datasets}). 
All the hyperparameter details (again tuned on the validation set) are presented in the supplementary. Table~\ref{table-entailmentprediction} presents the entailment generation results  for the baseline (sequence-to-sequence with attention, 3-ensemble, beam search) and the multi-task model which uses video captioning (shared decoder) on top of the baseline. A mixing ratio of $100:20$ alternate mini-batches of entailment generation and video captioning (resp.) works well.\footnote{Note that this many-to-one model prefers a different mixing ratio and learning rate than the many-to-one model for improving video captioning (Sec.~\ref{youtube-results}), because these hyperparameters depend on the primary task being improved, as also discussed in previous work~\cite{luong2015multi}.} The multi-task model achieves stat. significant ($p<0.01$) improvements over the baseline on all metrics, thus demonstrating that video captioning and entailment generation both mutually help each other.

\subsection{Human Evaluation}
\label{subsec-humanevaluation}

In addition to the automated evaluation metrics, we present pilot-scale human evaluations on the YouTube2Text (Table~\ref{table-baseandmultitaskresults}) and entailment generation (Table~\ref{table-entailmentprediction}) results. In each case, we compare our strongest baseline with our final multi-task model (M-to-M in case of video captioning and M-to-1 in case of entailment generation). We evaluate a random sample of $300$ generated captions (or entailed hypotheses) from the test set, across three human evaluators. We remove the model identity to anonymize the two models, and ask the human evaluators to choose the better model based on \emph{relevance} and \emph{coherence} (described in Sec.~\ref{subsec-evaluationmetrics}). As shown in Table~\ref{table-humanevaluation-video} and Table~\ref{table-humanevaluation-ent}, the multi-task models are always better than the strongest baseline for both video captioning and entailment generation, on both relevance and coherence, and with similar improvements (2-7\%) as the automatic metrics (shown in Table~\ref{table-baseandmultitaskresults}).


\begin{table}
\begin{center}
\begin{small}
\begin{tabular}{|l | c | c |}
\hline
 & Relevance & Coherence  \\
 \hline
 Not  Distinguishable & 70.7\% & 92.6\%  \\
 \hline
 SotA Baseline Wins & 12.3\% & 1.7\%\\
 Multi-Task Wins (M-to-M) & \textbf{17.0\%} & \textbf{5.7\%} \\
\hline
\end{tabular}
\vspace{-10pt}
\end{small}
\end{center}
\caption{Human evaluation on YouTube2Text video captioning.\vspace{-1pt}}
\label{table-humanevaluation-video}
\end{table}


\begin{table}
\begin{center}
\begin{small}
\begin{tabular}{|l | c | c | }
\hline
 &  Relevance & Coherence  \\
 \hline
 Not  Distinguishable &  84.6\% & 98.3\% \\
 \hline
 SotA Baseline Wins &  6.7\% & 0.7\% \\
 Multi-Task Wins (M-to-1) &  \textbf{8.7\%} & \textbf{1.0\%} \\
\hline
\end{tabular}
\vspace{-10pt}
\end{small}
\end{center}
\caption{Human evaluation on entailment generation.\vspace{-1pt}}
\label{table-humanevaluation-ent}
\end{table}

\begin{table}
\begin{center}
\begin{small}
\begin{tabular}{|p{0.31\textwidth-2\tabcolsep}|p{0.165\textwidth-2\tabcolsep}|}
\hline
Given Premise & Generated \;\;\;\;\;\;\; Entailment  \\
\hline
\hline
a man on stilts is playing a tuba for money on the boardwalk & a man is playing an instrument\\
\hline
a child that is dressed as spiderman is ringing the doorbell & a child is dressed as a superhero\\
\hline
several young people sit at a table playing poker & people are playing a game\\
\hline
a woman in a dress with two children &a woman is wearing a dress\\
\hline
a blue and silver monster truck making a huge jump over crushed cars & a truck is being driven\\

\hline
\end{tabular}
\vspace{-10pt}
\end{small}
\end{center}
\caption{Examples of our multi-task model's generated entailment hypotheses given a premise.}
\vspace{-15pt}
\label{table-entailmentexamples}
\end{table}


\subsection{Analysis}
Fig.~\ref{fig-msvdexamples} shows video captioning generation results on the YouTube2Text dataset where our final M-to-M multi-task model is compared with our strongest attention-based baseline model for three categories of videos: (a) complex examples where the multi-task model performs better than the baseline; (b) ambiguous examples (i.e., ground truth itself confusing) where multi-task model still correctly predicts one of the possible categories (c) complex examples where both models perform poorly. Overall, we find that the multi-task model generates captions that are better at both temporal action prediction and logical entailment (i.e., correct subset of full video premise) w.r.t. the ground truth captions. The supplementary also provides ablation examples of improvements by the 1-to-M video prediction based multi-task model alone, as well as by the M-to-1 entailment based multi-task model alone (over the baseline).

On analyzing the cases where the baseline is better than the final M-to-M multi-task model, we find that these are often scenarios where the multi-task model's caption is also correct but the baseline caption is a bit more specific, e.g., ``a man is holding a gun'' vs ``a man is shooting a gun''. 

Finally, Table~\ref{table-entailmentexamples} presents output examples of our entailment generation multi-task model (Sec.~\ref{subsec-entailmentprediction}), showing how the model accurately learns to produce logically implied subsets of the premise.

\section{Conclusion}
\label{sec-conclusion}
We presented a multimodal, multi-task learning approach to improve video captioning by incorporating temporally and logically directed knowledge via video prediction and entailment generation tasks.  We achieve the best reported results (and rank) on three datasets, based on multiple automatic and human evaluations. We also show mutual multi-task improvements on the new entailment generation task. In future work, we are applying our entailment-based multi-task paradigm to other directed language generation tasks such as image captioning and document summarization.

\section*{Acknowledgments}
We thank the anonymous reviewers for their helpful comments. This work was partially supported by a Google Faculty Research Award, an IBM Faculty Award, a Bloomberg Data Science Research Grant, and NVidia GPU awards.

\appendix
\vspace{15pt}
\noindent
{\fontsize{12}{12}\selectfont \textbf{Supplementary Material}} \par
\section{Experimental Setup}

\subsection{Datasets}

\subsubsection{Video Captioning Datasets}

\paragraph{YouTube2Text or MSVD}
The Microsoft Research Video Description Corpus (MSVD) or YouTube2Text~\cite{chen2011collecting} is used for our primary video captioning experiments.  It has $1970$ YouTube videos in the wild with many diverse captions in multiple languages for each video. Caption annotations to these videos are collected using Amazon Mechanical Turk (AMT). All our experiments use only English captions. On average, each video has $40$ captions, and the overall dataset has about $80,000$ unique video-caption pairs. The average clip duration is roughly $10$ seconds. We used the standard split as stated in~\newcite{venugopalan2015sequence}, i.e., $1200$ videos for training, $100$ videos for validation, and $670$ for testing. 

\paragraph{MSR-VTT}
MSR-VTT is a recent collection of  $10,000$ video clips of $41.2$ hours duration (i.e., average duration of $15$ seconds), which are annotated by AMT workers. It has  $200,000$ video clip-sentence pairs covering diverse content from a commercial video search engine. On average, each clip is annotated with $20$ natural language captions. We used the standard split as provided in~\cite{xu2016msr}, i.e.,  $6,513$ video clips for training, $497$ for validation, and $2,990$ for testing.

\paragraph{M-VAD}
M-VAD is  a movie description dataset with $49,000$ video clips collected from $92$ movies, with the average clip duration being $6$ seconds. Alignment of descriptions to video clips is done through an automatic  procedure using Descriptive Video Service (DVS) provided for the movies. Each video clip description has only $1$ or $2$ sentences, making  most evaluation metrics (except paraphrase-based METEOR) infeasible. Again, we used the standard train/val/test split as provided in~\newcite{torabi2015using}. 
\subsubsection{Video Prediction Dataset}
For our unsupervised video representation learning task, we use the UCF-101 action videos dataset~\cite{soomro2012ucf101}, which contains $13,320$ video clips of $101$ action categories and with an average clip length of $7.21$ seconds each.  
This dataset suits our video captioning task well because both contain short video clips of a single action or few actions, and hence using future frame prediction on UCF-101 helps learn more robust and context-aware video representations for our short clip video captioning task.
We use the standard split of $9,500$ videos for training (we don't need any validation set in our setup because we directly tune on the validation set of the video captioning task).

\subsection{Pre-trained Visual Frame Features}
For the three video captioning datasets (Youtube2Text, MSR-VTT, M-VAD) and the unsupervised video prediction dataset (UCF-101), we fix our sampling rate to $3fps$ to bring uniformity in the temporal representation of actions across all videos. These sampled frames are then converted into features using several state-of-the-art pre-trained models on ImageNet~\cite{deng2009imagenet} -- VGGNet~\cite{simonyan2014very}, GoogLeNet~\cite{szegedy2015going,ioffe2015batch}, and Inception-v4~\cite{szegedy2016inception}. For VGGNet, we use its $fc7$ layer features with dimension $4096$. For GoogLeNet and Inception-v4, we use the layer before the fully connected layer with dimensions $1024$ and $1536$, respectively.
We follow standard preprocessing and convert all the natural language descriptions to lower case and tokenize the sentences and remove punctuations.


\section{Training Details}
In all of our experiments, we tune all the model hyperparameters on validation (development) set of the corresponding dataset. We consider the following short hyperparameters ranges and tune lightly on: LSTM-RNN hidden state size - $\{256, 512, 1024\}$; learning rate in the range $[10^{-5}, 10^{-2}]$ with uniform intervals on a log-scale; weight initializations in the range $[-0.1,0.1]$ and mixing ratios in the range $1$:$[0.01,3]$ with uniform intervals on a log-scale. We use the following settings in all of our models (unless otherwise specified in a subsection below): we unroll video encoder/decoder LSTM-RNNs to $50$ time steps and language encoder/decoder LSTM-RNNs to $30$ time steps. We use a 1024-dimension LSTM-RNN hidden state size. We use $512$-dimension vectors to embed frame level visual features and word vectors. These embedding weights are learned during the training. We use the Adam optimizer~\cite{kingma2014adam} with default coefficients and a batch size of $32$. We apply a dropout with probability $0.5$ to the vertical connections of LSTM \cite{zaremba2014recurrent} to reduce overfitting. 

\subsection{Video Captioning on YouTube2Text}

\subsubsection{Baseline and Attention Models}
Our primary baseline model (Inception-v4, attention, ensemble) uses a learning rate of $0.0001$ and initializes all its weights with a uniform  distribution in the range $[-0.05,0.05]$. 

\subsubsection{Multi-Task with Video Prediction (1-to-M)}
In this  model, the video captioning and unsupervised video prediction tasks share their encoder LSTM-RNN weights and image embeddings in a one-to-many multi-task setting. We again use a learning rate of $0.0001$ and initialize all the learnable weights with a uniform distribution in the range $[-0.05,0.05]$. Two important hyperparameters tuned (on the validation set of captioning datasets) are the ratio of encoder vs decoder frames for video prediction on UCF-101 (where we found that $80\%$ of frames as input and $20\%$ for prediction performs best); and the mini-batch mixing ratio between the captioning and video prediction tasks (where we found $100:200$ works well).

\subsubsection{Multi-Task with Entailment Generation (M-to-1)}
In this  model, the video captioning and entailment generation tasks share their language decoder LSTM-RNN weights and word embeddings in a many-to-one multi-task setting. We again use a learning rate of $0.0001$. All the trainable weights are initialized with a uniform distribution in the range $[-0.08,0.08]$. We observe that a mixing ratio of $100:50$ (between the captioning and entailment generation tasks) alternating mini-batches works well here.

\subsubsection{Multi-Task with Video and Entailment Generation (M-to-M)}
In this many-to-many, three-task model, the video encoder is shared between the video captioning and unsupervised video prediction tasks, and the language decoder is shared between the video captioning and entailment generation tasks. We again use a learning rate of $0.0001$. All the trainable weights are initialized with a uniform distribution in the range $[-0.08,0.08]$. We found that a mixing ratio of $100:100:50$ alternative mini-batches of video captioning, unsupervised video prediction, and entailment generation works best.

\subsection{Video Captioning on MSR-VTT}
We also evaluate our many-to-many multi-task model on other video captioning datasets. For MSR-VTT, we train the model again using a learning rate of $0.0001$.  All the trainable weights are initialized with a uniform distribution in the range $[-0.05,0.05]$. We found that a mixing ratio of $100:20:20$ alternative mini-batches of video captioning, unsupervised video prediction, and entailment generation works best.

\begin{table*}[ht]
\begin{center}
\begin{small}
\begin{tabular}{|l|c|c|}
\hline
\multicolumn{3}{|c|}{\textsc{Multi-Task with Video Prediction (1-to-M)}}\\
\hline
\hline
Ground-truth Video Captions & Baseline &  Multi-Task (1-to-M) \\
\hline
a man drinks a glass of water &a man is eating something& a man is drinking something\\
 a man drinks something && \\
\hline
a man scores when playing basketball & a man is playing with a ball & a man is playing a basketball \\
young man dribbles and throws basketball &&\\
\hline
a person cuts a piece of blue paper &a man is playing with a board &a man is cutting a paper\\
a woman is cutting a paper in square by a scissor&&\\
\hline
a man is cutting meat with axe  &	a man is cooking &a man is cutting a piece of meat\\
a man is chopping a chicken && \\
\hline
a woman is slicing onions  &	a woman is slicing a vegetable &a woman is slicing an onion\\
a woman is chopping an onion && \\
\hline

a train is going down the track near a shore 	& a train is playing &a train is going on a track\\
a high speed train is running down the track&& \\
\hline
\hline
\multicolumn{3}{|c|}{\textsc{Multi-Task with Entailment Generation (M-to-1)}}\\
\hline
\hline
Ground-truth Video Captions & Baseline &  Multi-Task (M-to-1) \\
\hline
a cat is walking on the ground &a cat is playing with a cat&a cat is playing\\
a cat is sneaking through some grass && \\
\hline
a woman is applying eye liner & a woman is talking&a woman is doing makeup\\
a woman applies makeup to her eye brows&&\\
\hline
a baby tiger is playing & a tiger is playing with a tiger & a tiger is playing  \\
the tiger is playing &&\\
\hline
a man and woman are driving on a motorcycle  &a man is riding a song& a man is riding a motorcycle\\
 the man gave the woman a ride on the motorcycle && \\
\hline
a boy is walking on a treadmill  &a man is cleaning the floor&a man is exercising\\
a man exercising with a baby && \\
\hline
a puppy is playing on a sofa &a dog is playing with a dog&a puppy is playing\\
a puppy is running around on a sofa && \\
\hline
\end{tabular}
\end{small}
\end{center}
\caption{Examples showing cases where our one-to-many and many-to-one multi-task video-captioning models are better than the baseline.}
\label{table-multitaskentailmentandpreductionexamples}
\end{table*}

\subsection{Video Captioning on M-VAD}
For the M-VAD dataset, we use $512$ dimension hidden vectors for the LSTMs to reduce overfitting. We initialize the LSTM weights with a uniform distribution in the range $[-0.1,0.1]$ and all other weights with a uniform distribution in the range $[-0.05,0.05]$. We use a learning rate of $0.001$. We found a mixing ratio of $100:5:5$ alternative mini-batches of video captioning, unsupervised video prediction, and entailment generation works best.

\subsection{Entailment Generation}
Here, we use video captioning to in turn help improve entailment generation results. We use the same hyperparameters for both the baseline and the multi-task model (Sec.~5.3 and Table~4). We use a learning rate of $0.001$. All the trainable weights are initialized with a uniform distribution in the range $[-0.08,0.08]$. We found a mixing ratio of $100:20$ alternate mini-batches training of entailment generation and video captioning to perform best.

\section{Analysis}
In Sec.~5.5 of the main paper, we discussed examples comparing the generated captions of the final many-to-many multi-task model with those of the baseline. Here, we also separately compare our one-to-many (video prediction based) and many-to-one (entailment generation based) multi-task models with the baseline. As shown in Table~\ref{table-multitaskentailmentandpreductionexamples}, our one-to-many multi-task model better identifies the actions and objects in comparison to the baseline, because the video prediction task helps it learn better context-aware visual representations, e.g., ``a man is eating something'' vs. ``a man is drinking something'' and  ``a woman is slicing a vegetable'' vs. ``a woman is slicing an onion''.

On the other hand, the many-to-one multi-task (with entailment generation) seems to be stronger at generating a caption which is a logically-implied entailment of a ground-truth caption, e.g., ``a cat is playing with a cat'' vs. ``a cat is playing'' and ``a woman is talking'' vs ``a woman is doing makeup'' (see Table~\ref{table-multitaskentailmentandpreductionexamples}).

\bibliographystyle{acl_natbib}
\bibliography{citations}

\end{document}